\def\year{2020}\relax
\documentclass[letterpaper]{article} % DO NOT CHANGE THIS
\usepackage{aaai20}  % DO NOT CHANGE THIS
\usepackage{times}  % DO NOT CHANGE THIS
\usepackage{helvet} % DO NOT CHANGE THIS
\usepackage{courier}  % DO NOT CHANGE THIS
\usepackage[hyphens]{url}  % DO NOT CHANGE THIS
\usepackage{graphicx} % DO NOT CHANGE THIS
\usepackage{graphicx}  % DO NOT CHANGE THIS
\nocopyright %-- Your paper will not be published if you use this command
\usepackage{amsmath}
\usepackage{amssymb}
\usepackage{amsfonts}       % blackboard math symbols

\usepackage{tikz}
\usetikzlibrary{shapes,shapes.misc,backgrounds,decorations.pathreplacing,calc}

\usepackage{algpseudocode,algorithm}

\newcommand{\tikzmark}[1]{\tikz[overlay,remember picture] \node (#1) {};}

\makeatletter% Add a \tkizmark for each line so we can reference it later
    \algrenewcommand\alglinenumber[1]{\scriptsize#1\tikzmark{\arabic{ALG@line}}:}
\makeatother

\usepackage{subcaption}

\newcommand{\vc}[1]{\mathbf{#1}}
\newcommand{\prt}[1]{\overline{\mathbf{#1}}}

\usepackage[normalem]{ulem}

\title{Model-based Multi-Agent Reinforcement Learning\\with Cooperative Prioritized Sweeping}
\author{Eugenio Bargiacchi\textsuperscript{\rm 1}, Timothy Verstraeten\textsuperscript{\rm 1},\\
\Large \textbf{Diederik M. Roijers\textsuperscript{\rm 2}, Ann Now\'e\textsuperscript{\rm 1}}\\ % All authors must be in the same font size and format. Use \Large and \textbf to achieve this result when breaking a line
\textsuperscript{\rm 1}Vrije Universiteit Brussel\\ %If you have multiple authors and multiple affiliations
svalorzen@gmail.com % email address must be in roman text type, not monospace or sans serif
}
\begin{document}

\maketitle

\begin{abstract}
We present a new model-based reinforcement learning algorithm, \textit{Cooperative Prioritized Sweeping}, for efficient learning in multi-agent Markov decision processes. The algorithm allows for sample-efficient learning on large problems by exploiting a factorization to approximate the value function. Our approach only requires knowledge about the structure of the problem in the form of a dynamic decision network. %, and the structure of the approximated Q-function.
Using this information, our method learns a model of the environment and performs temporal difference updates which affect multiple joint states and actions at once. Batch updates are additionally performed which efficiently back-propagate knowledge throughout the factored Q-function. Our method outperforms the state-of-the-art algorithm sparse cooperative Q-learning algorithm, both on the well-known SysAdmin benchmark and randomized environments.
\end{abstract}

\noindent Consider a control problem where multiple agents must learn to cooperate to achieve a common goal in an environment with unknown and complex dynamics, such as robot soccer \cite{kok2003multi}, warehouse commissioning
\cite{claes2017decentralised}, and traffic light control \cite{wiering2000multi}. Such problems cannot be solved optimally in general even if the dynamics of the environment are known in advance, due to the size of both state and action spaces being exponential in the number of agents. Large environments also require increasingly large amounts of data to learn effectively, which are often impractical to obtain in real-life scenarios.

In this paper, we introduce an approximate method for fast, sample efficient reinforcement learning (RL) with a large set of cooperative agents, called \textit{cooperative prioritized sweeping} (CPS). Specifically, we generalize \textit{prioritized sweeping} (PS) \cite{ps} to linearly factored Q-functions \cite{cite:factoredlp,cite:ve}. We use model-based RL, which consists in learning a model of the environment, and using it to learn a
\textit{value function} and extract a policy for the agents. We represent this model using a \textit{dynamic decision network} and exploit the state-action dependency graph to efficiently construct and manage the priority queue. CPS performs updates similar to Q-Learning, but is able to back-propagate these changes through the Q-function by using the factored model. All updates are stored in a priority queue, such that the largest changes are back-propagated first in an anytime fashion.

\section{Problem Formulation}

A multi-agent Markov decision process (MMDP) is a tuple $\langle\vc{S}, \vc{A}, T, R, \gamma\rangle$, where $\vc{S}$ and $\vc{A}$ are the state and action spaces, $T$ is the transition function which describes the environment's dynamics, $R$ is the reward function associating rewards with joint state-action pairs and $\gamma \in [0, 1)$ is the discount factor representing the importance of future rewards. The joint action space is the Cartesian product of the action spaces of each agent, i.e., $\vc{A} = A_1 \times \dots \times A_K$. The state space is the Cartesian product of state factors, i.e., $\vc{S} = S_1 \times \dots \times S_N$.

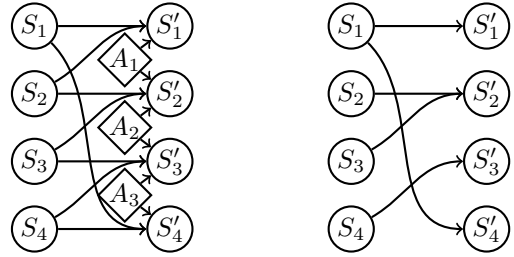
\begin{figure}[t]
\centering
\subcaptionbox{A simple DDN in compact form.\label{DDN}}[0.49\columnwidth]{
    \centering
    \begin{tikzpicture}[y=-1cm,scale=0.6]
        \begin{scope}[every node/.style={circle,thick,draw, minimum size =.6cm, inner sep=0pt}]
            \node (s1)  at (0,0.0) {$S_1$};
            \node (s2)  at (0,1.5) {$S_2$};
            \node (s3)  at (0,3.0) {$S_3$};
            \node (s4)  at (0,4.5) {$S_4$};
            \node (s11) at (3,0.0) {$S'_1$};
            \node (s22) at (3,1.5) {$S'_2$};
            \node (s33) at (3,3.0) {$S'_3$};
            \node (s44) at (3,4.5) {$S'_4$};
        \end{scope}

        \begin{scope}[every node/.style={diamond,thick,draw, minimum size =.6cm, inner sep=0pt}]
            \node (a1) at (2,0.75) {$A_1$};
            \node (a2) at (2,2.25) {$A_2$};
            \node (a3) at (2,3.75) {$A_3$};
        \end{scope}
        \draw [thick,->]
            (s1) edge (s11)
            (s1) edge[out=315, in=180](s44)

            (s2) edge (s22)
            (s2) edge[out=30, in=180](s11)

            (s3) edge[out=30, in=180](s22)
            (s3) edge (s33)

            (s4) edge[out=30, in=180](s33)
            (s4) edge (s44);

        \draw [thick,->]
            (a1) edge (s11)
            (a1) edge (s22)

            (a2) edge (s22)
            (a2) edge (s33)

            (a3) edge (s44)
            (a3) edge (s33);
    \end{tikzpicture}
}
\subcaptionbox{A DBN induced by a particular joint action.\label{DBN}}[0.49\columnwidth]{
    \centering
    \begin{tikzpicture}[y=-1cm,scale=0.6]
        \begin{scope}[every node/.style={circle,thick,draw, minimum size =.6cm, inner sep=0pt}]
            \node (s1)  at (0,0.0) {$S_1$};
            \node (s2)  at (0,1.5) {$S_2$};
            \node (s3)  at (0,3.0) {$S_3$};
            \node (s4)  at (0,4.5) {$S_4$};
            \node (s11) at (3,0.0) {$S'_1$};
            \node (s22) at (3,1.5) {$S'_2$};
            \node (s33) at (3,3.0) {$S'_3$};
            \node (s44) at (3,4.5) {$S'_4$};
        \end{scope}

        \draw [thick,->]
            (s1) edge (s11)
            (s1) edge[out=315, in=180](s44)

            (s2) edge (s22)

            (s3) edge[out=30, in=180](s22)

            (s4) edge[out=30, in=180](s33);
    \end{tikzpicture}
}
\caption{How to represent a multi-agent factored MDP.}
\label{DDNDBN}
\end{figure}

In an MMDP, the interactions between the state factors and agents are assumed to be \textit{sparse}. Therefore, we use a \textit{Dynamic Decision Network} (DDN) to describe them \cite{cite:ve}. A DDN can be graphically represented in a compact form as a directed graph where each edge marks the direct influence of one variable over another. Note that, while compact, such a representation can lose information, as in a DDN actions induce conditional dependencies which may be subsets of the edges represented. More specifically, an edge in the graph marks that a relationship exists for at least one action. For example, in Figure \ref{DDNDBN}, the state variable $S'_1$ may depend solely on $S_1$ when $A_1 = 0$, i.e., $P(S'_1\ |\ S_1, S_2, A_1 = 0) = P(S'_1\ |\ S_1, A_1 = 0)$, but may depend on both $S_1$ and $S_2$ when $A_1 = 1$. This information is part of the DDN itself--- but usually not represented graphically.

Using the DDN formulation, we can formulate the transition function of an MMDP as
\begin{equation}
T(\vc{s'}\ |\ \vc{s}, \vc{a}) = \prod_i T_i(s'_i\ |\ \prt{s}, \prt{a}),
\end{equation}
where $\prt{a}$ is the part of $\vc{a}$ on which $s'_i$ depends, and $\prt{s}$ is the part of $\vc{s}$ on which $s'_i$ depends given action $\prt{a}$. Therefore each joint action induces a \textit{Dynamic Bayesian Network} (DBN) out of the DDN (Figure \ref{DBN}).

We assume the reward function has the same structure as $T$, such that $R(\vc{s}, \vc{a}) = \sum_i R_i(\prt{s}, \prt{a})$ where each $R_i$ has the same domain $(\prt{S}, \prt{A})$ as $T_i$. Given this structure, we assume that rewards are sampled, both from the environment and from the model, as vectors with $N$ elements. This is useful when updating the factored Q-function (see Section \ref{qupdate}).

This representation is somewhat different from what some other multi-agent RL algorithms use (e.g., \cite{cite:scql}), which is to consider rewards on a per-agent basis, rather than on a
per-state factor basis. In other words, they consider reward samples to be vectors with $K$ entries, i.e., one per agent.
However, it is always possible to represent an agent-based reward function as a state-based one by simply adding one additional state factor per agent to convey the rewards.

\section{Cooperative Prioritized Sweeping}

To be able to learn in a factored MMDP, it is key to exploit the factored structure to learn an (approximate) sparse Q-function (Section \ref{qupdate}). For example, Sparse Cooperative Q-Learning (SCQL) \cite{cite:scql} learns a factorised Q-function on a per-agent or a per-edge (2-agent factor) basis. %is the state-of-the-art learning algorithm for MMDPs. It  %This method can leverage \textit{contextual specific independence}, meaning relationships which depend on the current state, by representing the Q-function as \textit{rules}.
However, %if we want to learn efficiently in the number of interactions with the environment,
the model-free approach with one-step look-ahead bootstrapping that SCQL uses is not very sample-efficient. %can be more flexible, but at the same time is less sample-efficient as it does not use fully the information it receives from the environment.
A straightforward idea to increase sample-efficiency would be to learn an approximate MMDP model (Section \ref{model}), and perform planning on this model to obtain an approximate sparse Q-function \cite{guestrin2003efficient,cite:ve}. %Similar to linear programs (LPs) for solving a single agent MDPs \cite{mdplinear},
The state-of-the-art planning algorithm employs an LP that computes the approximate value function that minimizes the max-norm distance to the optimal one \cite{cite:ve}. However, %this approach is only applicable when the full model is known in advance, and
because this requires solving a large scale LP at every time step as we learn the model, the computational requirements make such a model-based approach infeasible. % ONLY one LP is needed to solve an MMDP!!

In this section, we propose a new model-based approach for MMDPs that generalizes the idea of prioritized sweeping \cite{ps} to the multi-agent setting, by both using direct experience and simulated experience from an MMDP model (Section \ref{deltas}--\ref{queuesample}) to learn the sparse Q-function. Our algorithm---cooperative prioritized sweeping (Algorithm 1)---thus benefits from the sample-efficiency of a model-based approach, while keeping the computation time in check. % This enables us to find suitable trade-offs between computation time and sample-efficiency, by

\subsection{Learning the Model}\label{model}

Cooperative prioritized sweeping (CPS) approximates a model of the environment to determine how changes must be back-propagated through the value function after an update. Furthermore, this model is also used to generate new simulated experience that is sampled in batches and used to update the value function multiple times every time step. %In this paper, we assume the environment can be modeled as a factored MMDP as described above.

In order to learn the MMDP model, we assume that the DDN structure is fully described prior to learning. We use this structure for three purposes: to improve learning, to increase sample-efficiency, and to structure the CPS queue so that it contains parent sets rather than complete states (see Section \ref{queueupdate}). Given the structure of the MMDP, we learn the transition dynamics by maintaining estimates of the proportion that a state $s'_i$ is reached after taking a local joint action $\prt{a}$ in a local joint state $\prt{s}$ of its parents in the DDN \cite{genps}:
\begin{equation}
T_i(s'_i\ |\ \prt{s}, \prt{a}) = \frac{N_{\prt{s},\prt{a},s'_i} + N^0_{\prt{s},\prt{a},s'_i}}
{\sum_{s_i'' \in S_i} N_{\prt{s},\prt{a},s_i''} + N^0_{\prt{s},\prt{a},s_i''}}
\end{equation}
where $N^0_{\prt{s},\prt{a}, s_i''}$ are the priors on the model before interaction begins. The reward function can be learned accordingly.

\subsection{One-step Q-Function Updates}\label{qupdate}

CPS learns the Q-function of the MMDP in a factorized manner. This means that it approximates the true Q-function as a linear sum of smaller components:
\begin{equation}
Q(\vc{s}, \vc{a}) \approx \hat{Q}(\vc{s}, \vc{a}) = \sum_x \hat{Q}_x(\prt{s}, \prt{a})
\label{eq:Q_approx}
\end{equation}
We assume that the domains of each $\hat{Q}_x$ are obtained by back-propagating some subset of the components in $\vc{S}$ (we call these \textit{basis domains}). This is the same process that is used to obtain $\hat{Q}$ from a suitably factored value function $\hat{V}$ made out of basis functions in approximate MMDP planning \cite{cite:ve}.
This assumption is important in order to correctly back-propagate rewards. For example, in Figure \ref{DDN}, selecting variable $S'_1$ as a basis results in a corresponding domain of its $\hat{Q}$ component of $S_1 \land S_2 \land
A_1$. Selecting both $S'_1 \land S'_2$ results in a corresponding domain of $S_1 \land S_2
\land S_3 \land A_1 \land A_2$. The set of basis domains can be chosen arbitrarily---although some
choices tend to perform better than others. Defining larger and more domains leads to better accuracy, but at the cost of higher computation times for joint action selection. For example, CPS uses variable elimination to select the optimal action, which is known to be exponential in the induced width of the coordination graph of the factored Q-function (after conditioning on the current state $\vc{s}$), which in turn increases with the size and number of the domains in the factorization. Constructing a $\hat{Q}$ made up of a single factor with the entire state space $\vc{S}$ as its basis degenerates the algorithm to single-agent PS, as it would make $\hat{Q} = Q$.

After each interaction with the environment, we obtain an experience tuple, $\langle \vc{s}, \vc{a}, \vc{s'}, \vc{r} \rangle$, which CPS uses to update the factored Q-function. Since CPS takes a model-based approach, the ideal update
would first use the data to update the model (both $T$ and $R$), and then use the model to update
the Q-function:

\begin{equation}
\hat{Q}(\vc{s}, \vc{a}) = R(\vc{s}, \vc{a}) + \gamma \sum_\vc{s'} T(\vc{s'}\ |\ \vc{s}, \vc{a}) \hat{V}(\vc{s'})
\end{equation}
% FIXME: Use pluseq? Or reduce font size? (can do up to 6.5 point)

Indeed, this operation, using a $\hat{V}$ factored with the relevant basis domains, is not only possible but
also relatively cheap \cite{cite:factoredlp}:  each $\hat{Q}_x$ component can be updated separately, while ensuring that $\hat{Q}$ fully reflects the changes of the just updated model.
However, back-propagating the changes made to $\hat{Q}$ to $\hat{V}$ is not straightforward; to do so, an algorithm would need to solve an expensive LP \cite{cite:factoredlp}. This operation makes performing batches of updates to the Q-function computationally infeasible, and would prevent the algorithm from leveraging the learnt MMDP model.

Because full backups are not feasible for the factored Q-functions, CPS employs local TD updates:

\begin{equation}
\hat{Q}(\vc{s}, \vc{a}) = \hat{Q}(\vc{s}, \vc{a}) + \alpha \left ( R(\vc{s}, \vc{a}) + \gamma \max_{\vc{a'}} \hat{Q}(\vc{s'}, \vc{a'}) - \hat{Q}(\vc{s}, \vc{a}) \right )
\end{equation}

% We'll leave this for another time...
% Note that this can probably be slightly improved since $R(\vc{s}, \vc{a})$ is not the sampled
% reward but our learned reward function, so we could leave within the $\alpha$ only the update of the
% future value.

To perform this update, CPS requires the optimal joint action, $\vc{a'^\star}$, for the next joint state $\vc{s'}$, and thus needs to maximize $\hat{Q}$ for this joint state. This can
be done efficiently using \textit{Variable Elimination} (VE).\footnote{We note that VE is an exact algorithm for coordination graphs that can be replaced by approximate algorithms \cite{kok2005using,marinescu2005and,liu2012belief} if need be.} Once the optimal full joint action
$\vc{a'^\star}$ for $\vc{s'}$ has been found, we can then split the update of the Q-function per component:

\begin{equation}
\hat{Q}_x(\prt{s}, \prt{a})= \hat{Q}_x(\prt{s}, \prt{a}) + \alpha \left( R_x(\prt{s}, \prt{a}) + \gamma \hat{Q}_x(\prt{s'}, \prt{a'^\star}) - \hat{Q}_x(\prt{s}, \prt{a}) \right )
\label{eq:Q_x_update}
\end{equation}

The only thing left is to compute $R_x(\prt{s}, \prt{a})$, as its domain will likely be different from any of the $R_i$ functions in the learnt MMDP model, due to the back-propagation of the basis domains. Hence, for each $S_i$ in the original basis domain of $\hat{Q}_x$, we compute $R_x$ as a weighted sum of their respective $R_i$, where the weights are inversely proportional to how many $\hat{Q}_y$ have $S_i$ in their original basis domain:

\begin{equation}
R_x(\prt{s}, \prt{a}) = \sum_i I(S_i \in Basis_x(\prt{S})) \frac{R_i(\prt{\prt{s}}, \prt{\prt{a}})}{N_{\hat{Q}}(Basis(S_i))}
\end{equation}

where $I$ is the indicator function and $N_{\hat{Q}}(Basis(S_i))$ is the number of $\hat{Q}_y$ factors ($\hat{Q}_x$ included) which have $S_i$ in their basis
domain. We also indicate with $\prt{\prt{s}}, \prt{\prt{a}}$ the parts of $\prt{s}, \prt{a}$ required by $R_i$. This works because we are assuming that $R$ has the
same DDN structure as $T$, and due to the back-propagation, if $R_x$ had $S_i$ in its basis domain,
then $\prt{s}$ and $\prt{a}$ must contain the parents of $s_i$ which are required by $R_i$.

These updates are similar to those performed in SCQL \cite{cite:scql,kok2005using}. However, CPS determines the shape of the Q-function and how it partitions the rewards using the basis domains, while SCQL does this in terms of agents and their direct neighbours.

\subsection{Temporal Differences for Batch Q-Function Updates}\label{deltas}

In order to perform batch updates on the Q-function to back-propagate useful values, CPS
computes a series of temporal differences (TD) which represent how much the Q-function has changed due to the last
update. This process is different from the single-agent PS algorithm \cite{ps,genps} in two key
aspects: how we define the TD, and how CPS computes it.

Single-agent PS computes a single TD for the state $s$ that is being updated. In our case, we update each $\hat{Q}_x$ independently, resulting in a series of
$\Delta_x$, defined as
\begin{equation}
\Delta_x = R_x(\prt{s}, \prt{a}) + \gamma \hat{Q}_x(\prt{s'}, \prt{a'^\star}) - \hat{Q}_x(\prt{s}, \prt{a}).
\end{equation}
Given the factored Q-function (Equation~\ref{eq:Q_approx}) and factor update rule (Equation~\ref{eq:Q_x_update}), we could reconstruct a full TD by simply summing them together, i.e., $\Delta = \sum_x \Delta_x$. However, given the structure of our MMDP, computing a single TD for the whole state $\vc{s}$ is wasteful: given the structure of our MMDP, each $\Delta_x$ of $\hat{Q}_x(\prt{s}, \prt{a})$ not only reflects changes for the single state $\vc{s}$, but all states that contain $\prt{s}$. In the same way, the change in value of $\hat{Q}_x(\prt{s}, \prt{a})$ not only affects the value of all possible parents of $\vc{s}$, but of all possible parents of $\prt{s}$.
An interesting consequence of this fact is that our definition of a TD changes from the single-agent PS algorithm, where the TD is computed with respect to the change in the value function for the updated state $s$. This is because there is no reason to perform additional updates if the actual value function for the state is not modified. In our case, this is different, as any single $\hat{Q}_x(\prt{s}, \prt{a})$ value might contribute to $\hat{V}(\vc{s})$ in some state where $\prt{s} \subseteq \vc{s}$---although actually obtaining this information is not practical. Instead, we assume that the change is indeed modifying $\hat{V}$, and compute the TDs directly on the change that CPS performs without the need to actually check $\hat{V}$ or perform any additional maximizations. As these TDs are computed without the knowledge of how much each $\Delta_x$ actually affects $\hat{V}$, this means that CPS tends to overestimate changes in order to not miss any.

Given these considerations, our goal is to obtain a series of TDs which:
\begin{enumerate}
\item Reflect the actual changes we have applied to the Q-function, without passing through $\hat{V}$.
\item Do not lose the information about how much each part of the state space was changed.
\item Are suitable for back-propagation in order to select new $(\vc{s}, \vc{a})$ pairs to update.
\end{enumerate}

The $\Delta_x$ satisfy the first two constraints, but not the third, as back-propagating their domain would mismatch the domain of any $\hat{Q}_x$. Therefore, we transform them into a set of $N$ elements $\Delta_i$, one for each factor of the state-space
\begin{equation}
\Delta_i = I(i \in x_s) \frac{\Delta_x}{|x_s|}
\end{equation}
where $x_s$ refers to the state domain part of $\hat{Q}_x$. In this format, these can be used in the back-propagation step (see Section \ref{queueupdate}). Note that the $\Delta_i$ are only used to sample from the queue, and not in the actual $\hat{Q}$ updates.

\subsection{Updating the Queue}\label{queueupdate}

CPS maintains a priority queue containing the changes it has done to the
Q-function, where the priority is equal to the magnitude of the change. The key idea is to
back-propagate these changes during batch updates to the Q-values of the parents of each $s_i$. Since CPS must do this given the limited time between interactions with the environment, the updates should be done from the highest priority to the lowest until the time runs out.

CPS uses the $\Delta_i$ to update the queue. Note that we go a single step backwards in time, such that our $\vc{s}$ is now considered a
\textit{future state}. Then, for each $s_i$, CPS iterates over all the possible values of its
parents $Parents(s_i) = (\prt{s''}, \prt{a''})$ as determined by the DDN of the problem.

The algorithm can now compute the priority $p$ of these parents as $p = T_i(s_i | Parents(s_i)) \cdot \Delta_i$,
i.e., the probability of observing this transition times the estimated amount of change of that particular
component. Note that we do not include $\gamma$ in the priority since it would not affect the
relative ordering of the queue. If $p$ is greater than a threshold value $\theta$, it is added as a new entry to the queue, or increase its value by $p$ if such an
entry already exists. The entry will contain the parents of $s'_i$ (and their
values), together with the computed priority $p$.

\subsection{Sampling from the Queue}\label{queuesample}

CPS needs to sample from the queue to perform the batch updates required to increase the accuracy of the Q-function. In single-agent PS, this operation is relatively straightforward: a state-action pair is sampled, and a new Q update is done.
However, in CPS, the queue only contains partial state-action pairs, while a complete state-action pair is necessary to sample a new state and reward from the model and select the next optimal joint action needed to perform a full Q-function update. Thus, CPS extracts multiple compatible entries from the priority queue, and combines them into a complete state-action pair.

While the optimal choice would be the set with the highest combined priority, finding this set
is equivalent to the knapsack problem, which is known to be NP-hard.
A good heuristic would be to traverse the priority queue in order, greedily extracting
the first $(\prt{s}, \prt{a})$ entry from the queue, and
then greedily selecting all entries which are compatible with both it and previously selected entries,
where compatible means that the same parents $(s_i, a_i)$ must have the same values in all selected
entries. Unfortunately, this option is also not viable, as it would require examining all entries
in the queue, which would be too computationally expensive to do repeatedly. However, by ignoring the order of priority, this strategy can be implemented very efficiently with the help of \textit{radix trees} for bookkeeping, which significantly reduces the number of comparisons needed. Thus, CPS still pops the highest priority entry from the queue,
but randomizes the subsequent search order. In this way, we ensure that every compatible entry in the queue has a chance of being selected, and that all updates are performed in the limit.

Once the set of entries has been finalized, CPS removes them from the priority queue, and proceeds to use them to update the factored Q-function. If some elements local state-action $(s_i, a_i)$ required to construct a full joint action are not in the selected set, CPS supplements these by uniformly sampling them from their respective domain. Another option would be to iterate over every possible combination of
the missing values and do an update for each combination, but this operation would too expensive, as the number of updates would increase exponentially with the number of unassigned values. % \textcolor{red}{(Timo comment: we could remove the previous sentence. if the reviewers point this out, it's an easy answer. ofcourse, only if we are in need of space.)}
Finally, using the newly obtained $(\vc{s}$, $\vc{a})$ pair, CPS uses the model to sample a new state $\vc{s'}$ and rewards $\vc{r}$, and updates $\hat{Q}$ as described in Section \ref{qupdate}.

\begin{center}
\hrulefill\\\vspace{3pt}
\textbf{Algorithm 1:} Cooperative Prioritized Sweeping\\
% NOTE: The vspace below is to make the algorithm's title symmetric.
% I could not find another way to do it. It does not impact anything else.
\vspace{-2pt}\hrulefill
\end{center}
\begin{algorithmic}[1]
% FIXME: Removed notes since they could not fit
\While{True}
    \State Select action $\vc{a}$ given state $\vc{s}$ following policy $\pi$
    \State $(\vc{s'}, \vc{r}) \leftarrow$ Execute action $\vc{a}$ in state $\vc{s}$
    \State Update $T, R$ using $(\vc{s}, \vc{a}, \vc{s'}, \vc{r})$
% \AddNote[blue]{3}{4}{Section \ref{model}}
    \State $\vc{a'^\star} \leftarrow$ Use variable elimination for $\vc{s'}$
    \State Update $\hat{Q}(\vc{s}, \vc{a})$ using $(\vc{s}, \vc{a}, \vc{s'}, \vc{a'^\star})$
% \AddNote[blue]{5}{6}{Section \ref{qupdate}}
    \For{every state factor $i$}
        \State $\Delta_i \leftarrow$ Compute from TDs of $\hat{Q}$
%         \AddNoteSingle[blue]{8}{Section \ref{deltas}}
        \For{every pair $\prt{s''}, \prt{a''}$ that is parent of $s_i$} % FIXME: I removed this since it is redundant -> where $T(s'_i = s_i\ |\ \prt{s''}, \prt{a''}) > 0$}
            \State $p \leftarrow \Delta_i \cdot T(s'_i = s_i\ |\ \prt{s''}, \prt{a''})$
            \If{$p > \theta$}
                \State Add to Queue $\prt{s''}, \prt{a''}$ with priority $p$
            \EndIf
        \EndFor
    \EndFor
% \AddNote[blue]{9}{13}{Section \ref{queueupdate}}
    \For{number of batch updates}
        \State $(\prt{s}, \prt{a}) \leftarrow$ Pop Queue max
        \For{every pair $(\prt{s''}, \prt{a''})$ in Queue, randomly}
            \If{$(\prt{s}, \prt{a})$ and $(\prt{s''}, \prt{a''})$ are compatible}
                \State Append $(\prt{s''}, \prt{a''})$ to $(\prt{s}, \prt{a})$
                \State Remove $(\prt{s''}, \prt{a''})$ from Queue
            \EndIf
        \EndFor
        \State $(\vc{s}, \vc{a}) \leftarrow$ Fill missing values of $(\prt{s}, \prt{a})$ randomly
        \State $(\vc{s'}, \vc{r}) \leftarrow$ Sample from $T, R$ using $\vc{s}$ and $\vc{a}$
        \State Update $\hat{Q}$
        \State Update Queue with all $\Delta_i$\;
    \EndFor
% \AddNote[blue]{16}{24}{Section \ref{queuesample}}
% \AddNote[blue]{25}{26}{Repeat the process of Sections\\\ref{qupdate}, \ref{deltas}, \ref{queueupdate}}
\EndWhile
\end{algorithmic}

\section{Experiments}

\begin{figure*}[t!]
\centering
\subcaptionbox{Regret over time w.r.t.\ the best possible approximation (LP) in the shared ring control setting.\label{shared}}{
    \centering
    \includegraphics[width=0.4\textwidth]{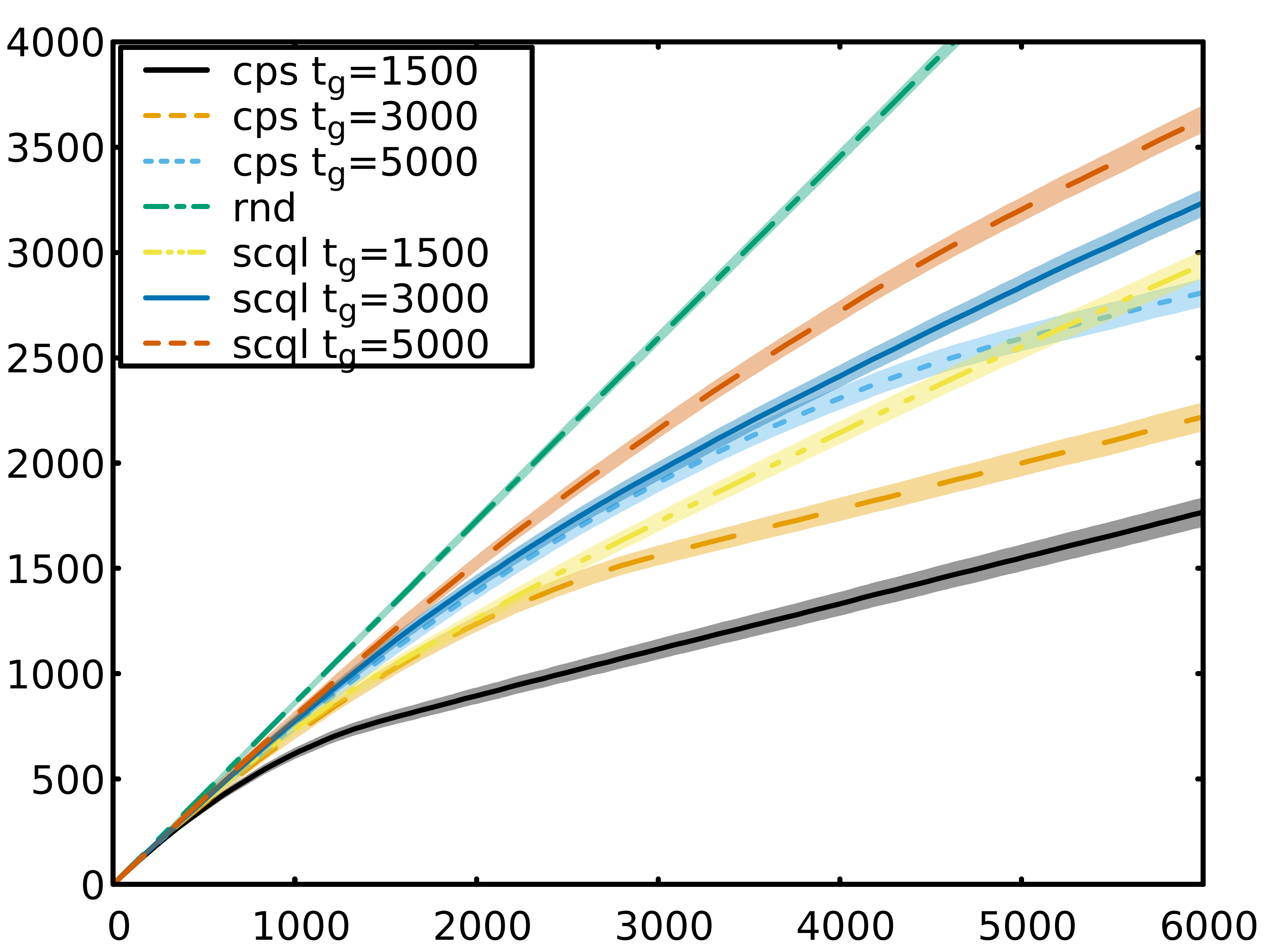}
}
\subcaptionbox{Regret over time w.r.t.\ CPS $t_g=250$ in the ring setting, with 300 agents.}{
    \centering
    \includegraphics[width=0.4\textwidth]{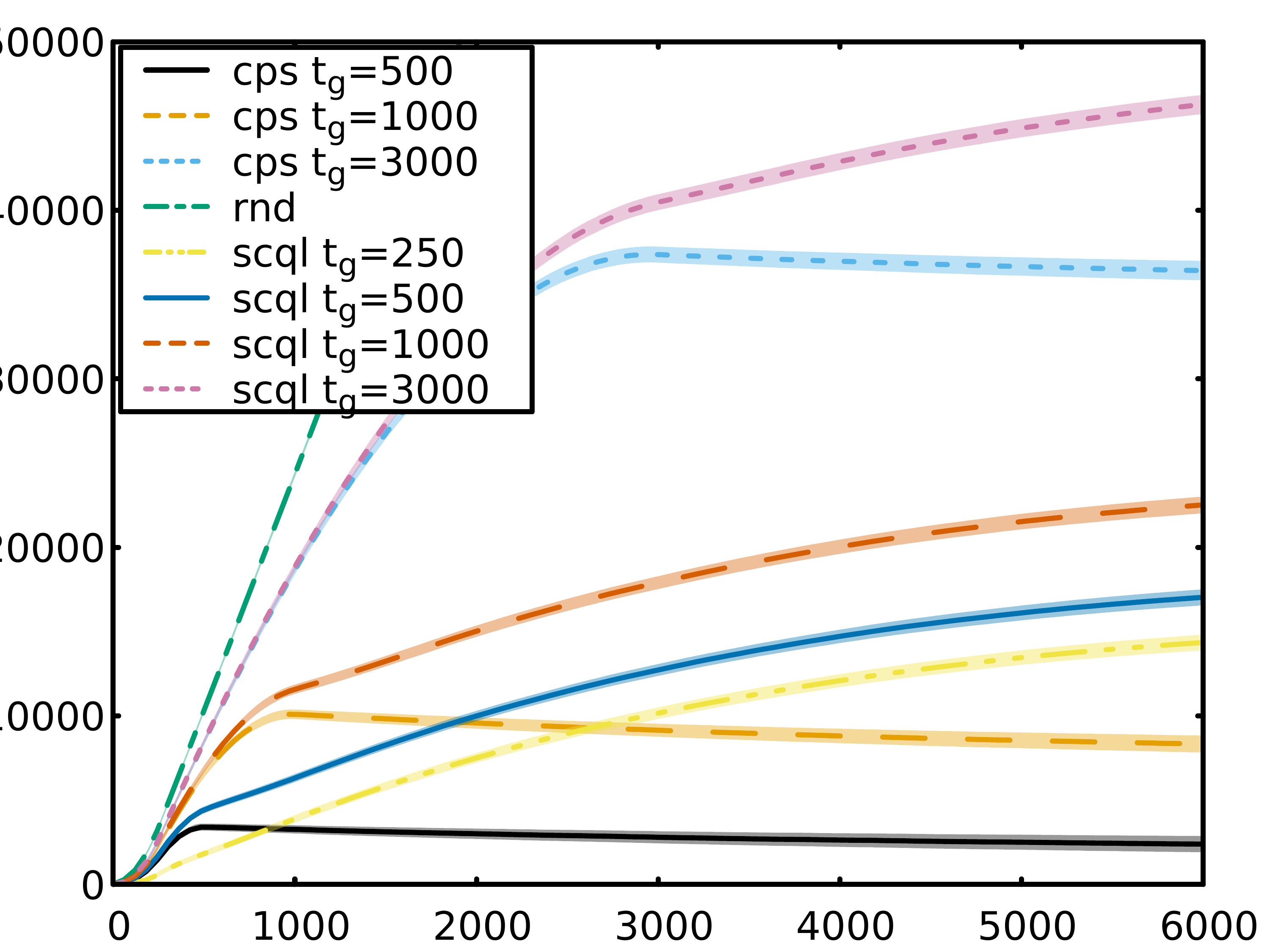}
}
\subcaptionbox{Regret over time w.r.t.\ the best possible approximation (LP) in the random setting with 3 agents.\label{random-small}}{
    \centering
    \includegraphics[width=0.4\textwidth]{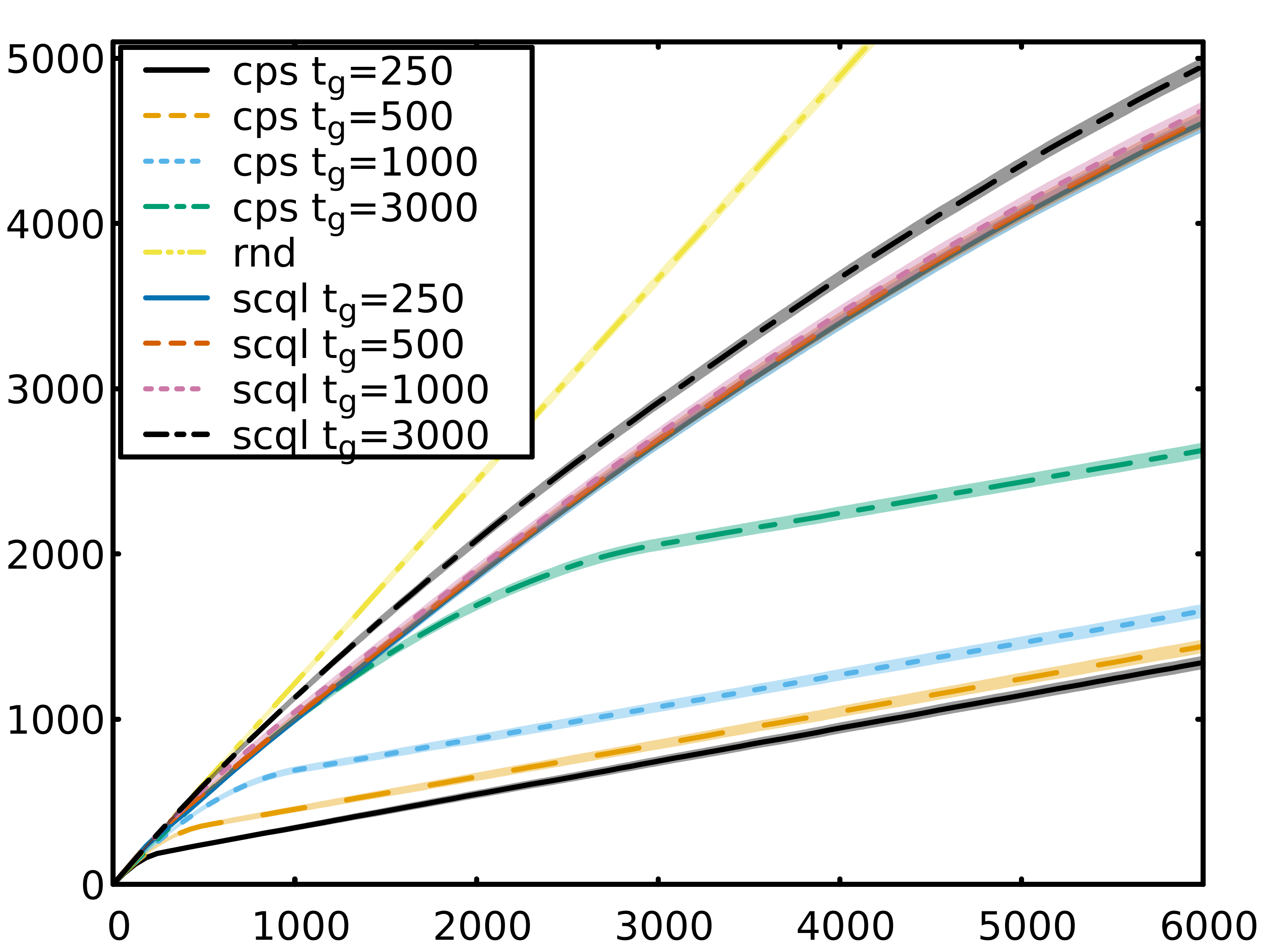}
}
\subcaptionbox{Regret over time w.r.t.\ CPS $t_g=250$ in the torus setting, with 100 agents.}{
    \centering
    \includegraphics[width=0.4\textwidth]{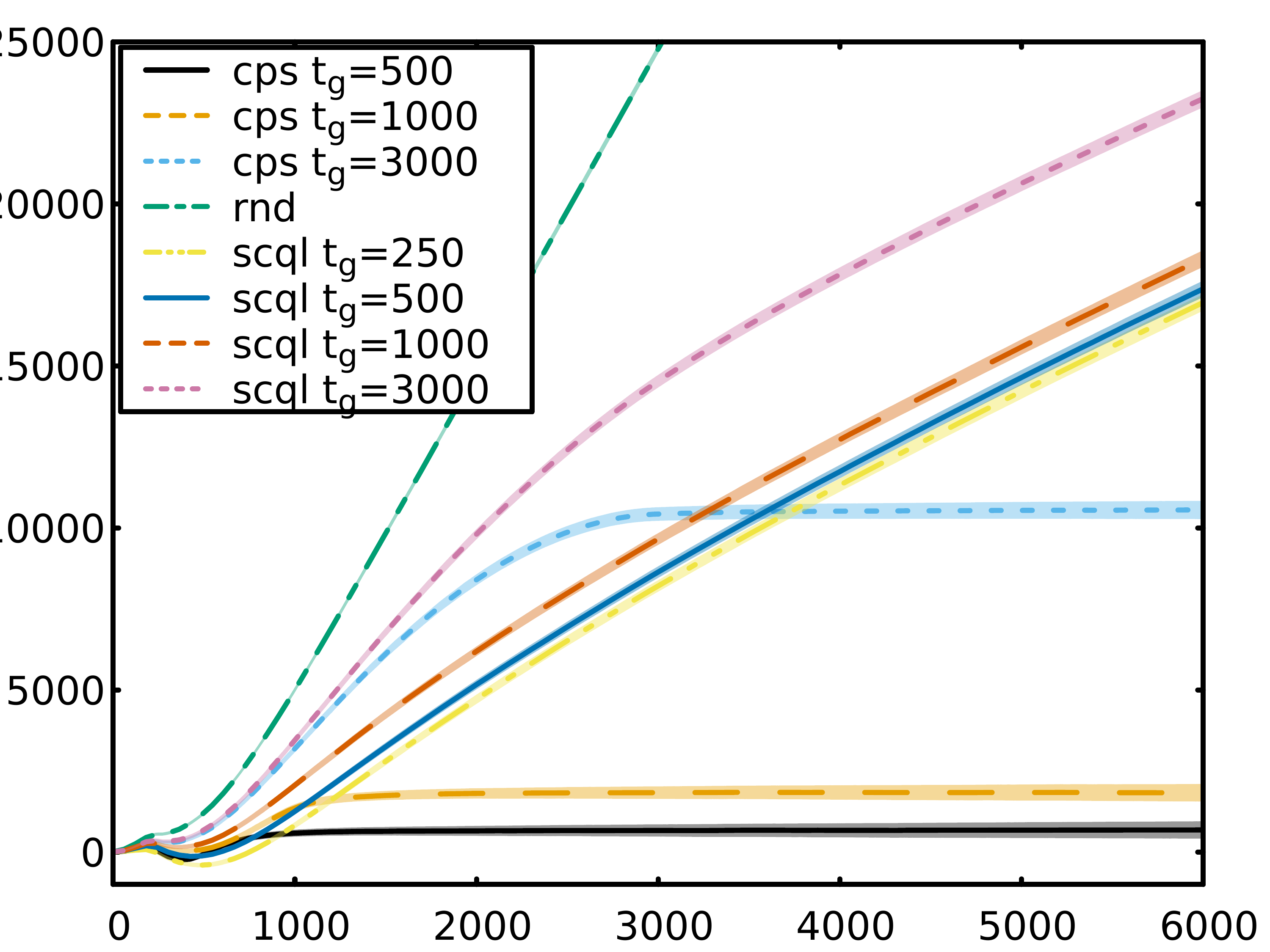}
}
\subcaptionbox{Regret over time w.r.t.\ a batch size of $50$ in the torus setting. $t_g$ here is $1000$ for all runs.\label{batch}}{
    \centering
    \includegraphics[width=0.4\textwidth]{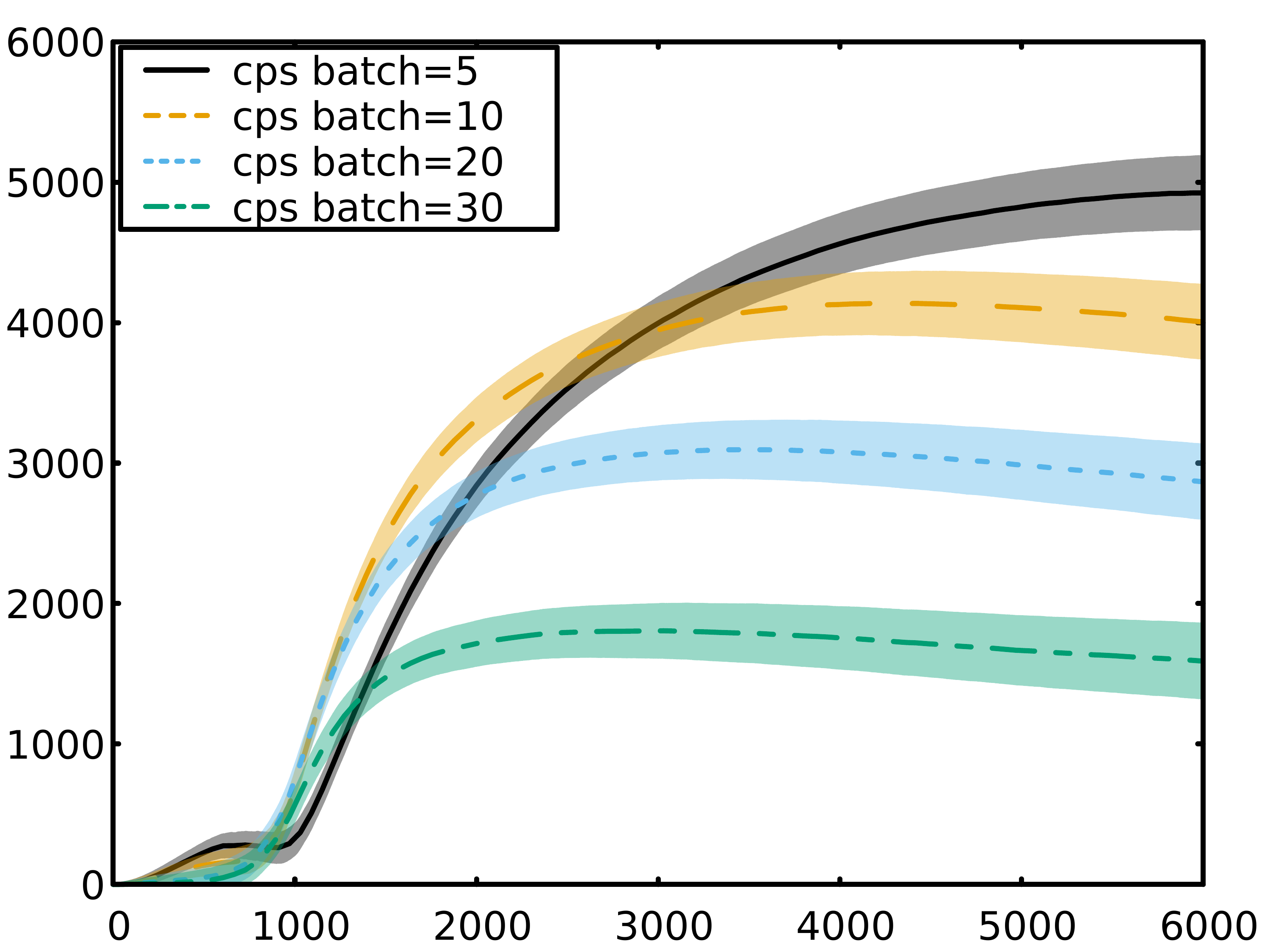}
}
\subcaptionbox{Regret over time w.r.t.\ CPS $t_g=250$ in the random setting, with 15 agents.\label{random-big}}{
    \centering
    \includegraphics[width=0.4\textwidth]{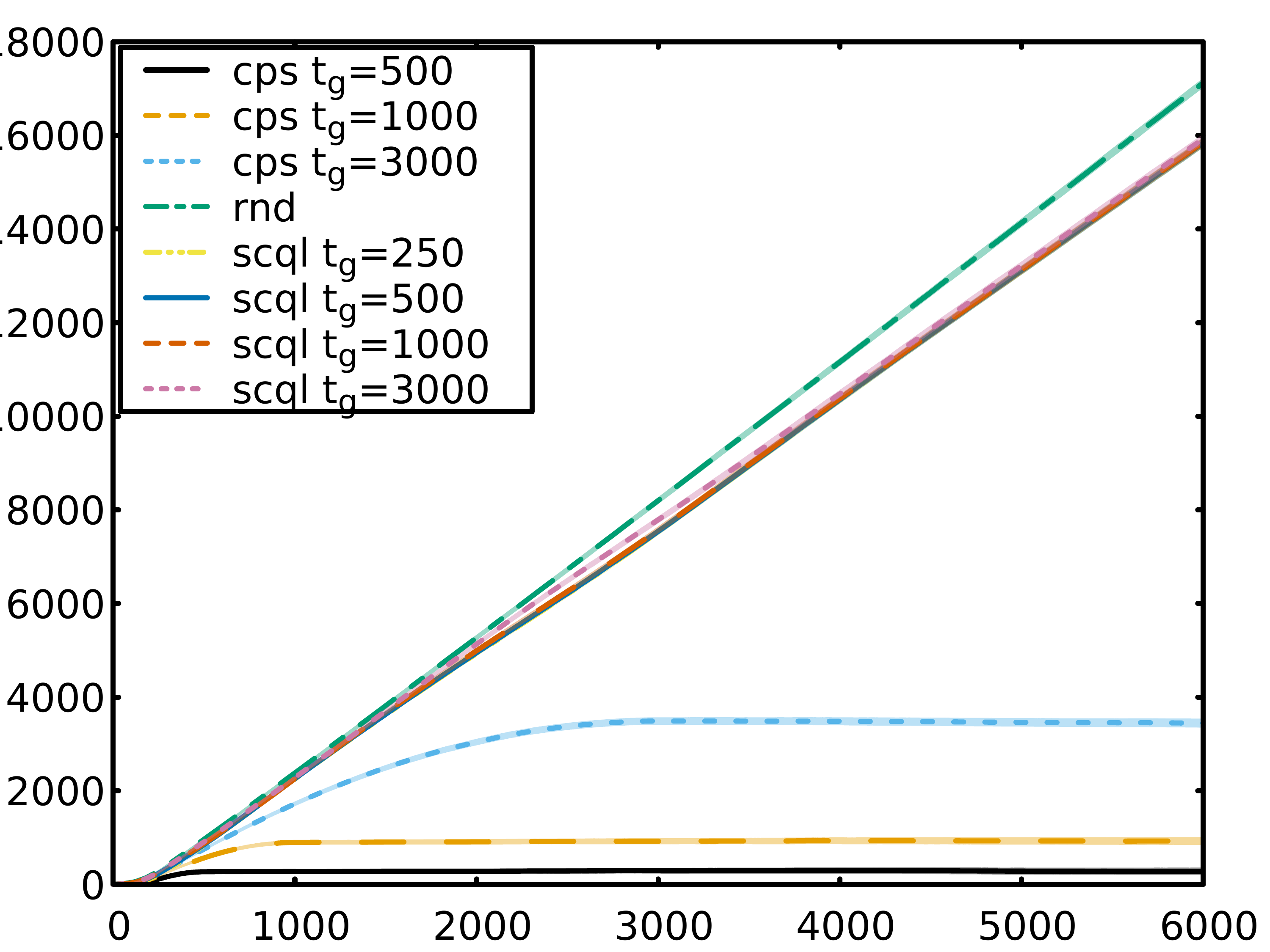}
}
\caption{Mean and standard deviation of cumulative regret over time for different methods in each setting (lower is better). Note that in all plots but \ref{shared} and \ref{random-small} regret may go down since it is computed w.r.t.\ a specific algorithm, as the LP upper bound is too expensive to compute.}
\label{plots}
\end{figure*}

To test the performance of CPS, we evaluate it empirically on the SysAdmin problem \cite{cite:ve} and randomly generated MMDPs.

In the SysAdmin setting, a series of interconnected machines randomly obtain jobs that they must complete. Over time, the machines may fail or die, and this in turn affects the probability of failures for adjacent machines. Each machine is associated with an agent, which can either do nothing or reboot its machine. Rebooting a machine resets its state, but loses all work. Each completed job receives a reward of 1 for its associated agent. We use three different topologies: a ring with 300 agents, a torus with 100 agents in a 10x10 grid and a shared control setting with a ring of 12 machines where each is controlled by its two adjacent agents. In this last setting, a machine reboots with a probability of 0.15 if only one agent issues the rebooting command, and with a probability of 1 if both do. We only use 12 agents so that it is feasible to use a factored LP \cite{cite:ve} to obtain the best possible approximation and compute the true regret. We select each machine's state and load pair as the basis domains, which is equivalent to the ``single'' basis reported in \cite{cite:ve}.

Additionally, we evaluate the performance of CPS on randomly generated MMDPs. The DDNs for these problems were generated by connecting each $S'_i$ with $S_i$, plus a random number of state and action factors (max 3) locally close to $i$. The transition tables were filled by uniformly sampling transition vectors. We generated sparse reward functions containing random values in $[-1, 0, 1]$ for the parents of random factors $S_i$, which are selected independently with probability $0.3$. We select as basis domains all pairs of adjacent state factors $(S_i, S_{i+1})$. We show results for two randomly generated MMDPS, one with 4 state factors and 3 agents, and another with 20 state factors and 15 agents. Both environments had to be kept relatively small, as on one hand the LP approach was very slow in these less structured environments, while on the other SCQL had trouble handling the large number of rules required to represent the Q-function (in the millions) with more than 15 agents.

We compare our results against 3 benchmarks: a random policy as a naive approach, the factored LP planning method on the ground truth MMDP model as the upper bound \cite{cite:ve}, and Sparse Cooperative Q-Learning (SCQL) \cite{cite:scql}. SCQL runs used optimistic initialization to improve its exploration, with all its Q-function values set to $5.0$, as without it SCQL was not able to learn correctly. Both SCQL and CPS use a constant learning rate of $0.3$, and an $\varepsilon$-greedy policy with $\varepsilon$ linearly decreasing from $0.9$ to $0$. We tested different time steps at which the exploration probability $\varepsilon$ must reach 0, reported as parameter $t_g$ in Figure~\ref{plots}. CPS, SCQL and the LP approach all use the same Q-function factorization. CPS has no prior knowledge about the transition and reward functions at the beginning of each run. All results were averaged over 100 runs. Each experiment ran on a single core of an AMD FX-8350 CPU.

CPS consistently outperforms SCQL in all settings, and is able to obtain good policies with very few samples: 250 time steps are enough to learn a close-to-optimal policy in an environment with $2^{300}$ actions and $9^{300}$ states. At the same time, exploring for a longer time leads to slightly better policies.

Figure \ref{batch} shows the relationship between regret and batch size in CPS in the SysAdmin torus setting. As the batch size decreases, regret increases. However, smaller batch sizes still converge to the same policy over time.
Because of the limited exploration time, the learnt model is imperfect. This results in slightly sub-optimal decisions inversely correlated with the batch size, as batch updates are sampled from the learnt model. Therefore, lower batch sizes show decreasing regret in the later time steps.

\section{Related Work}

In general, exact planning in factored multi-agent settings is unfeasible for larger numbers of agents, for all but problems with very few agents \cite{valuefunctiondependencies}. Some sub-classes of decentralized versions of such problems exist that do permit an exact factorization of the value function \cite{becker2004solving,nair2005networked,witwicki2010influence}. However, the cost of decentralization is still arbitrarily large, and planning in these specialized models is still expensive. Therefore, for model-based reinforcement learning it is better to directly approximate a factored sparse Q-function, as this obviates the need for the additional over-simplifying model restrictions that using such models would entail.

In this paper, we make use of variable elimination \cite{cite:ve} to select the optimal joint action given w.r.t.\ an approximate value function. Such joint action selection can also be done approximately, e.g., via max-plus \cite{kok2005using}, AND/OR tree search \cite{marinescu2005and}, or variational methods \cite{liu2012belief}. We note that this is orthogonal to our contributions and can be added easily if the complexity of the chosen graphical structure of the value function requires it.

Finally, recent advances have been made in deep multi-agent reinforcement learning \cite{gupta2017cooperative,foerster2018counterfactual,deepmarlsurvey}. The most related method to ours from this line of research is QMIX \cite{rashid2018qmix} which---to our knowledge---is the only deep MARL method to exploit some type of factorization. However, this method is limited to Q-functions that are monotonic in its constitution components, and thus not yet able to handle arbitrary factorizations of the Q-function. It is our hope that the contributions presented in this paper will also contribute to future deep MARL methods that can handle any factorization, especially in light of the fact that it has been observed that the sparse factored representation for the value functions we employ is equivalent to a single layer neural network with features corresponding to its basis functions \cite{cite:ve}.

\section{Discussion and Conclusions}

We have presented a new model-based reinforcement learning algorithm, Cooperative Prioritized Sweeping, which exploits the structure of a coordination graph in an MMDP to both learn a model of the environment and update the Q-function in a sample-efficient manner. The algorithm maintains a priority queue containing partial state-action pairs, representing the parts of the Q-function which are most beneficial to update. The queue is sampled in batches between interactions with the environment, and the model is used to sample possible new state and reward pairs, which are then used to update the Q-function. CPS can learn good policies using significantly less data than the previous state-of-the-art, resulting in significantly less regret over time.

%CPS does not yet support contextual-specific independence \cite{cite:verules, cite:scql}, as we rely on dense representations for the factors in $T$, $R$ and $\hat{Q}$. However, it is possible to represent these functions using rules \cite{cite:verules}. However, using a rule-based representation might have a significant performance impact on updating and sampling from the priority queue.

Currently, CPS uses a naive $\varepsilon$-greedy exploration strategy. In future work, we aim to improve exploration; instead of randomly sampling a completely random joint-action when exploring, we would like to select the joint-action that maximizes the collection of information from each $T$ and $R$ factor, similar to previous work in multi-agent bandits \cite{mauce}. Such an approach could further reduce the sample complexity of CPS. We will also implement support for context-specific independence using rules \cite{cite:verules}. Finally, we aim to use the techniques presented in this paper as a basis for constructing a deep MARL that can fully support arbitrary factored Q-functions.

%\section{Acknowledgements}
% The acknowledgments section, if included, appears after the main body of text and is headed ``Acknowledgments." This section includes acknowledgments of help from associates and colleagues, credits to sponsoring agencies, financial support, and permission to publish. Please acknowledge other contributors, grant support, and so forth, in this section. Do not put acknowledgments in a footnote on the first page. If your grant agency requires acknowledgment of the grant on page 1, limit the footnote to the required statement, and put the remaining acknowledgments at the back. Please try to limit acknowledgments to no more than three sentences.

\bibliography{bibl}
\bibliographystyle{aaai}
\end{document}